\documentclass[letterpaper]{article} 
\usepackage{aaai25}  
\usepackage{times}  
\usepackage{helvet}  
\usepackage{courier}  
\usepackage[hyphens]{url}  
\usepackage{graphicx} 
\usepackage{amssymb}

\usepackage[table,xcdraw]{xcolor}
\usepackage{natbib}  
\usepackage{caption} 
\usepackage{algorithm}
\usepackage{algorithmic}
\usepackage{adjustbox}
\usepackage{multirow}
\usepackage{amsmath}
\usepackage{booktabs}
\usepackage{amsmath}
\usepackage[table,xcdraw]{xcolor}

\usepackage{pifont} 

\newcommand{\cmark}{\ding{51}}%
\newcommand{\xmark}{\ding{55}}%
\usepackage{newfloat}
\usepackage{listings}

\DeclareCaptionStyle{ruled}{labelfont=normalfont,labelsep=colon,strut=off} 
\floatstyle{ruled}
\newfloat{listing}{tb}{lst}{}
\floatname{listing}{Listing}
\usepackage{newfloat}
\usepackage{listings}
\DeclareCaptionStyle{ruled}{labelfont=normalfont,labelsep=colon,strut=off} 
\floatstyle{ruled}
\newfloat{listing}{tb}{lst}{}
\floatname{listing}{Listing}
\title{Cross-Modal Few-Shot Learning with Second-Order Neural Ordinary Differential Equations}
\author {
    Yi Zhang\textsuperscript{\rm 1,\rm 2}\equalcontrib,
    Chun-Wun Cheng\textsuperscript{\rm 3}\equalcontrib,
    Junyi He\textsuperscript{\rm 2},
    Zhihai He\textsuperscript{\rm 2,4{\dag}},
    Carola-Bibiane Schönlieb\textsuperscript{\rm 3},
    Yuyan Chen\textsuperscript{\rm 5},
    Angelica I Aviles-Rivero\textsuperscript{\rm 6}\thanks{Corresponding Authors}
}
\affiliations {
    \textsuperscript{\rm 1}Harbin Institute of Technology, China\\
    \textsuperscript{\rm 2}Department of Electrical and
Electronic Engineering, Southern University of Science and Technology, China\\
    \textsuperscript{\rm 3}Department of Applied Mathematics and Theoretical Physics, University of Cambridge, UK\\
    \textsuperscript{\rm 4}Pengcheng Laboratory, China\\
    \textsuperscript{\rm 5}Shanghai Key Laboratory of Data Science, School of Computer Science, Fudan University, China\\
    \textsuperscript{\rm 6}Yau Mathematical Sciences Center, Tsinghua University, China\\

    zhangyi2021@mail.sustech.edu.cn, cwc56@cam.ac.uk, hejy2021@mail.sustech.edu.cn, hezh@sustech.edu.cn,
    cbs31@cam.ac.uk, chenyuyan21@m.fudan.edu.cn,
    aviles-rivero@tsinghua.edu.cn
}

\usepackage{bibentry}

\begin{document}

\maketitle

\begin{abstract}
We introduce SONO, a novel method leveraging Second-Order Neural Ordinary Differential Equations (Second-Order NODEs) to enhance cross-modal few-shot learning. By employing a simple yet effective architecture consisting of a Second-Order NODEs model paired with a cross-modal classifier, SONO addresses the significant challenge of overfitting, which is common in few-shot scenarios due to limited training examples. Our second-order approach can approximate a broader class of functions, enhancing the model's expressive power and feature generalization capabilities. We initialize our cross-modal classifier with text embeddings derived from class-relevant prompts, streamlining training efficiency by avoiding the need for frequent text encoder processing. Additionally, we utilize text-based image augmentation, exploiting CLIP’s robust image-text correlation to enrich training data significantly. Extensive experiments across multiple datasets demonstrate that SONO outperforms existing state-of-the-art methods in few-shot learning performance.
\end{abstract}

%

\section{Introduction}
\label{sec:intro}
Contrastive vision-language pre-training has revolutionized multimodal machine learning, establishing a new framework for integrating visual and textual data \cite{li2022blip,jia2021scaling}. This approach has rapidly gained traction, influencing a wide range of visual tasks such as semantic segmentation, object detection, image captioning, and classification~\cite{yuan2021multimodal,rao2022denseclip,wang2022cris,wang2023efficient}.
CLIP~\cite{radford2021learning}, one of the most recognized vision-language models, has garnered widespread attention for its simplicity and effectiveness. Trained on large-scale image-text pairs, CLIP creates a unified embedding space by aligning visual and textual modalities, enabling strong zero-shot performance across various downstream tasks~\cite{zhang2022tip,zhu2023notAPE}. However, despite its competitive performance, CLIP's pre-trained nature limits its adaptability to unseen domains. To enhance CLIP's performance in few-shot settings, several works have focused on fine-tuning by adding learnable modules on top of the frozen CLIP model to better handle new semantic domains. 

\begin{figure}[t!]
\begin{center}
\centerline{\includegraphics[width=1\linewidth]{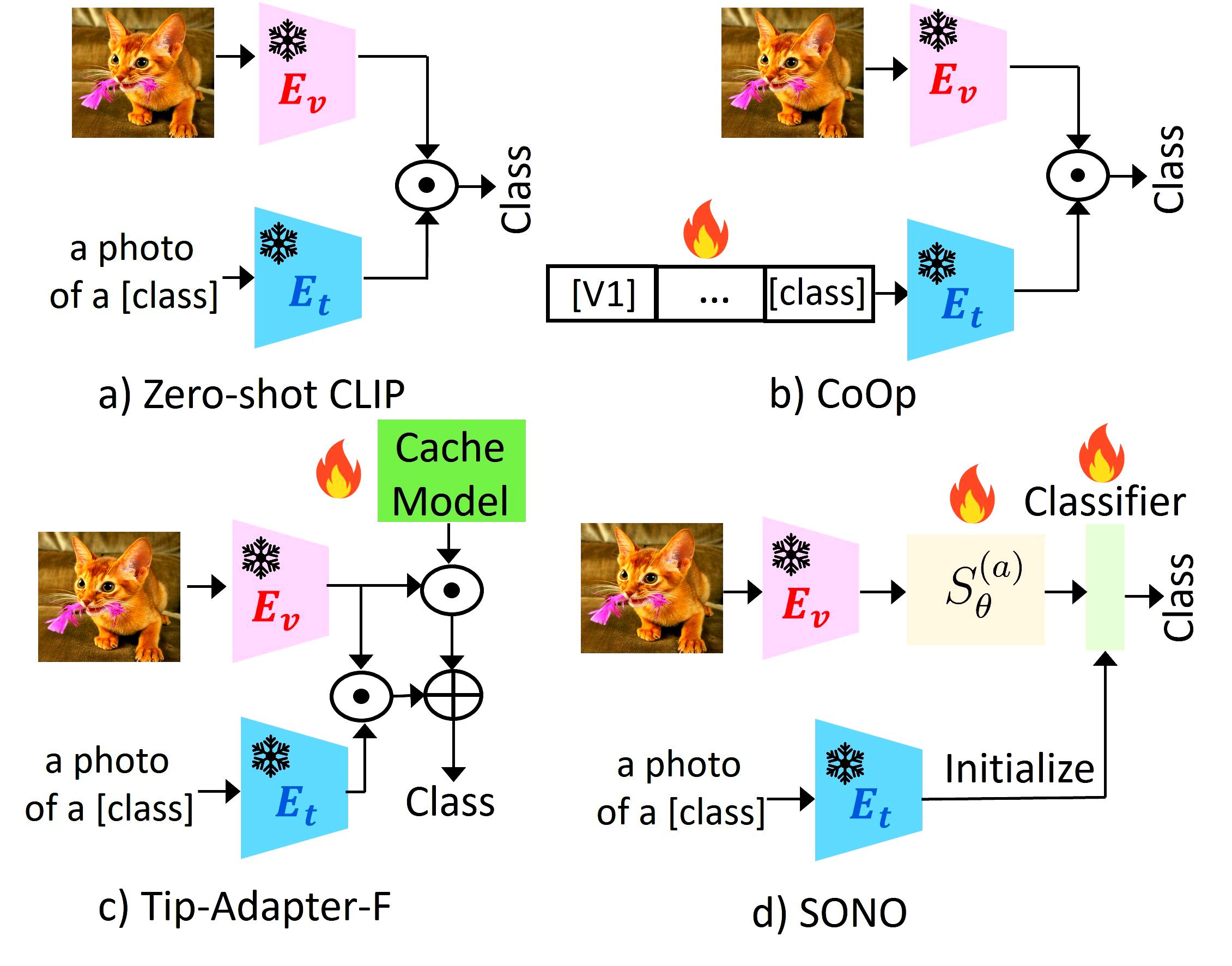}}

\caption{Comparison between (a) Zero-shot CLIP~\cite{radford2021learning}, (b) CoOp~\cite{zhou2022learning}, (c) Tip-Adapter-F~\cite{zhang2022tip}, and (d) our proposed SONO, where $S^{(a)}_\theta$ represents the Second-Order NODE model.
}
\label{fig:rect}
\end{center}
\end{figure}

Existing CLIP fine-tuning methods can be broadly categorized into two groups: 1) input-level prompting methods, such as CoOp~\cite{zhou2022learning}, CoCoOp~\cite{zhou2022conditional}, and PLOT~\cite{chen2023plot}, and 2) feature-level adapting methods, like CLIP-Adapter~\cite{gao2021clip}, Tip-Adapter-F~\cite{zhang2022tip}, and GraphAdapter~\cite{li2024graphadapter}. Input-level prompting methods use learnable prompts before CLIP's text encoder to distill task-specific knowledge, while feature-level adapting methods apply residual-style adapters after CLIP's encoders, as shown in Figure~\ref{fig:rect}. However, prompting methods such as CoOp show limited few-shot accuracy and require additional training time and computational resources, while adapter-based methods like Tip-Adapter-F can be inefficient due to the large caches and extensive parameter tuning. This leads us to ask: \textit{Is it possible to achieve strong few-shot performance by adding only efficient learnable modules, making fine-tuning both effective and efficient?}

To address these challenges, we propose a novel method called SONO, which is simple yet efficient. 
SONO introduces Second-Order Neural Ordinary Differential Equations (Second-Order NODEs)  as a powerful approach for cross-modal few-shot learning, with a particular emphasis on feature optimization.
Few-shot classification challenges models with very limited examples per class, leading to a high risk of overfitting in traditional neural networks. Neural Ordinary Differential Equations (NODEs) 
address this by modeling data transformations continuously, rather than through discrete layers. This continuous modeling enables smoother, more generalized feature transformations that are less likely to overfit. NODEs typically use fewer parameters, reducing the risk of memorizing data and further preventing overfitting. Additionally, NODEs utilize ODE solvers for dynamic time step adjustment during training, enhancing the precision of feature optimization and overall model performance.

Despite the strengths of NODEs, few-shot classification requires models with high expressive power. NODEs, limited by their inability to approximate a broad class of functions, can struggle in these tasks. This is because their ODE flows do not intersect, restricting their capability to capture complex patterns. Augmenting these to Second-Order NODEs, which are universal approximators \cite{kidger2022neural}, resolves this issue by broadening the range of functions they can model and enhancing their representational capacity. This allows for faster training and convergence. With these improvements, Second-Order NODEs maintain continuous transformation and offer a robust solution for few-shot classification, promising improved performance.

\textbf{Contributions.} Our contributions could be summarized as follows: 1) We introduce a novel method called SONO, which employs a simple architecture consisting of a Second-Order Neural ODE model and a cross-modal classifier. 2) We use the Second-Order NODEs for feature optimization. The cross-modal classifier is initialized with text embeddings derived from prompts containing class names. This initialization approach makes the classifier functions similarly to prompt tuning in CoOp, but it eliminates the need to process data through the text encoder in every training iteration, making our method more efficient. 3) Additionally, we adopt a strategy of using text as images for data augmentation. Given CLIP's powerful image-text correlation capabilities and the ease of obtaining text descriptions for each class, this strategy proves to be highly effective. For each class, we select several prompts with the highest cosine similarities to the available labeled training images, using them as data augmentation for the training samples. 4) Our extensive experiments demonstrate that our proposed SONO significantly improves few-shot classification and domain generalization performance, surpassing state-of-the-art methods by a substantial margin.

\section{Related Work}

\label{sec:related work}

\begin{figure*}[ht]

\begin{center}
\centerline{\includegraphics[width=\linewidth]{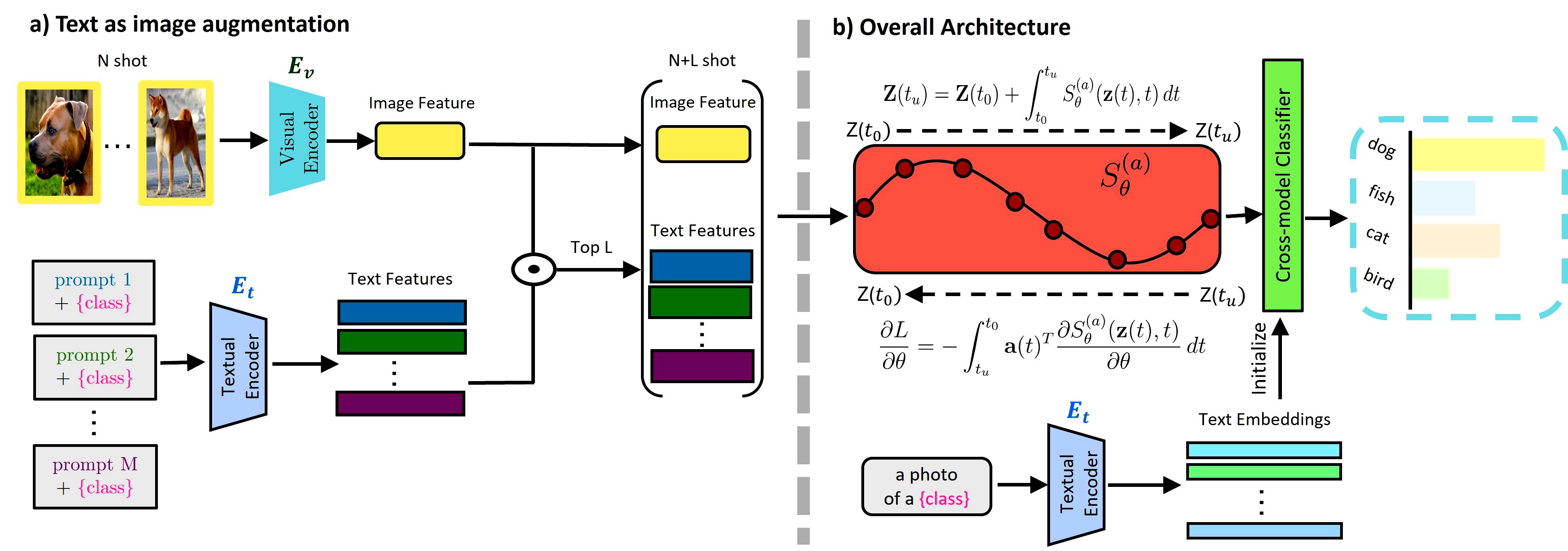}}

\caption{\textbf{An overview of our method for $K$-class $N$-shot classification}. Subfigure (a) illustrates the text-as-image data augmentation process. Subfigure (b) presents the overall architecture of our proposed SONO, consisting of a Second-Order NODEs model $S^{(a)}_{\theta}$ and a cross-modal classifier, which is initialized with text embeddings derived from prompts containing class labels.}
\label{fig:overview}
\end{center}

\end{figure*}

\textbf{Vision-Language Models (VLMs).}
VLMs have become a central focus in multimodal research due to their capacity to integrate visual and textual data into a shared representation space~\cite{yang2022vision, su2019vl, desai2021virtex}. These models can be broadly classified based on their pre-training objectives into contrastive learning~\cite{radford2021learning, jia2021scaling}, generative modeling~\cite{yu2022coca, singh2022flava}, and alignment objectives~\cite{li2022grounded, yao2022detclip}. Contrastive learning, exemplified by models such as CLIP~\cite{radford2021learning} and ALIGN~\cite{jia2021scaling}, leverages extensive image-text datasets to align visual and textual embeddings, resulting in impressive performance on zero-shot learning and other open-world tasks. In this work, we utilize CLIP \cite{radford2021learning} as the foundation for our approach.

\textbf{Fine-tuning for VLMs.}
Recent efforts to adapt pre-trained VLMs to downstream tasks focus on two main strategies: \textit{Prompt Tuning} and \textit{Feature Adaptation} \cite{zhang2024vision}. Prompt tuning optimizes the input prompts, either textual or visual, to better align with downstream tasks while keeping most VLM parameters fixed. Notable examples include CoOp~\cite{zhou2022learning}, which optimizes context words for each class, and CoCoOp~\cite{zhou2022conditional}, which conditions prompts on individual images to improve generalization. Other approaches, such as TPT~\cite{shu2022tpt}, adapt prompts dynamically during test-time to better handle distribution shifts. Feature adaptation involves introducing lightweight adapters to refine the representations learned by VLMs. For instance, CLIP-Adapter~\cite{gao2021clip} adds simple linear layers, and Tip-Adapter~\cite{zhang2022tip} offers a training-free approach by directly using few-shot embeddings as a cache model, enabling efficient adaptation. Additionally, TaskRes~\cite{yu2023task} adapts the text-based classifier to better exploit the existing knowledge in the pre-trained VLM, while GraphAdapter~\cite{li2024graphadapter} leverages task-specific structures and relationships within the data for more specialized adaptation. Besides these two major approaches, methods like CuPL~\cite{pratt2023does}, which uses large language models to generate more effective prompts, and CALIP~\cite{guo2022calip}, which introduces parameter-free attention mechanisms, offer additional innovations. This work diverges from both prompt tuning and feature adaptation, presenting a unique approach using NODEs for efficient cross-modal few-shot learning.

\textbf{Neural ODEs.} Neural Ordinary Differential Equations (NODEs) \cite{chen2018neural} extend the concept of Residual Networks (ResNets) \cite{he2016deep} by considering the limit where the discretization step approaches zero. This approach naturally leads to the formulation of an Ordinary Differential Equations (ODEs), which can be optimized using black-box ODE solvers. The continuous-depth nature of NODEs makes them exceptionally well-suited for learning and modeling the unknown dynamics of complex systems, which are often difficult to describe analytically. It has been applied to Vision-Language Reasoning \cite{zhang2024node}.
However, numerous dynamical systems encountered in scientific research, including Newton's equations of motion and various oscillatory systems, are primarily governed by Second-Order ODEs. \cite{massaroli2020dissecting} demonstrated that higher-order systems exhibit greater parameter efficiency. Furthermore, \cite{norcliffe2020second} conducted a comprehensive study on Second-Order behaviors, concluding that these systems perform better. Applications of these models have been explored in fields such as medical segmentation \cite{cheng2023continuous} and the acceleration of diffusion models \cite{ordonezmissing}. Despite these promising outcomes, there is a conspicuous gap in the research concerning the application of VLMs. In this study, we focus on leveraging Second-Order NODEs to fine-tune VLMs and enhance feature optimization.

\section{Method}
\label{sec:method}
\subsection{Background}
\textbf{CLIP.} The CLIP model~\cite{radford2021learning} excels in visual tasks by mapping image and text into a joint embedding space using contrastive learning on large-scale image-text pairs. CLIP's encoders, $\{E_t, E_v\}$, include a text encoder $E_t$ (usually a Transformer \cite{vaswani2017attention}) and an image encoder $E_v$ (typically a ResNet \cite{he2016deep} or ViT \cite{dosovitskiy2020image}). In a zero-shot classification setting with $N$ classes, given a test image $x_{test}$, the visual feature $f_{v} = E_v(x_{test})$ is extracted, and $N$ text features $f_{t_i} = E_t(\{\pi; y_i\})$ are generated, where $y_i$ is appended to a prompt $\pi$, (e.g.,``a photo of a''). The probability of $x_{test}$ belonging to $y_i$ is calculated as:
\begin{equation}
\label{eq-clip}
   p(y = y_i|x_{test})=\frac{\exp \left( \mathrm{sim}\left(f_{t_i},f_{v} \right) /\tau \right)}{\sum\nolimits_{t'}{\exp \left( \mathrm{sim}\left( f_{t'},f_{v} \right) /\tau \right)}}, 
\end{equation}
where $\tau$ is the softmax temperature and $\mathrm{sim}(\cdot, \cdot)$ denotes cosine similarity.

\subsection{Cross-modal Few-shot Learning with Second-Order Neural Ordinary Differential Equations (SONO)}

\textbf{Method Overview. } 
In Figure~\ref{fig:overview}, we provide an overview of our proposed SONO. Figure~\ref{fig:overview}(a) illustrates the Text-as-Image Augmentation process. Leveraging CLIP's strong image-text correlation capabilities, we use text for data augmentation. For a $K$-class $N$-shot few-shot learning problem, given a class $k$ and a list of prompts containing class labels $\Psi$ collected from CLIP~\cite{radford2021learning} and CuPL~\cite{pratt2023does}, we select $L$ prompts from the $M$ available for each class using cosine similarity as the augmentation for class $k$.
The overall architecture of SONO is shown in Figure~\ref{fig:overview}(b). SONO consists of a Second-Order Neural ODE model for feature optimization and a cross-modal classifier for classification. Given a feature sample from the augmented training features from (a), it is first input into the Second-Order Neural ODE model to refine the feature, then the refined feature is fed into the cross-modal classifier for the final prediction. The classifier is initialized with text embeddings derived from prompts containing class labels.

\subsection{Second-Order Neural ODEs for Better Features \label{sec:sonode}}

This section details the mathematical foundation of Second-Order Neural ODEs for Better Features, providing a detailed exploration of the underlying principles that enable these models to effectively refine features. 

In our framework, we define our model as follows:
\begin{equation}
 \left\{\begin{array}{l}\mathbf{x}^{\prime\prime}(t)=S_\theta^{(a)}\left(\mathbf{x}(t), \mathbf{x}^{\prime}(t), t \right) \\ \mathbf{x}(t_0) = x_0 ,~~~ \mathbf{x}^{\prime}(t_0)= g_\theta(x(t_0)),
 \end{array}\right.
   \label{eq:important}
\end{equation}
where $S_\theta^{(a)}$ is typically a neural network parameterized by $\theta$, $\mathbf{x}(t_0)$ and $\mathbf{x}^{\prime}(t_0)$ are the two initial conditions. We first explore the computational aspects of~\eqref{eq:important}. This model employs ODE solvers in both forward pass and backpropagation. However, ODE solvers only allow the use of first-order ordinary differential equations. Therefore, we need to transform our model into a system of first order NODEs first. To represent~\eqref{eq:important} as a system of first-order NODEs, we introduce the state vector $\mathbf{z}(t)$: $\mathbf{z}(t) = [\mathbf{x}(t), \mathbf{x}^{\prime}(t)]^T$ 
$\implies$ $\mathbf{z}^{\prime}(t)=[x'(t), S_\theta^{(a)}\left(\mathbf{x}(t), \mathbf{x}^{\prime}(t), t\right)]^T = S_\theta^{(v)}(\mathbf{z},t,\theta_{f})$
with the initial condition: $\mathbf{z}(t_0)=[X_0,g(X_0, \theta_{g})]^T$.
We can now apply the ODE solvers. Given an initial feature \( \mathbf{z}(t_0) \), our objective is to refine this feature progressively until we obtain the final optimal feature, denoted as \( \mathbf{z}(t_u) \).  The feature \( \mathbf{z}(t_0) \) was derived from the Text-as-Image Augmentation process, whereas the feature \( \mathbf{z}(t_u) \) was subsequently updated by the ODE solver at the final time step. In the context of the ODE solver, the fourth-order Runge-Kutta method (RK4) method was selected due to its well-known convergence rate and enhanced stability properties. Additionally, the maximum time step for the ODE solvers was set to 1000, indicating that the maximum value of \( u \) is 1000. The ODE solver was employed in both the forward pass and backpropagation processes.

\noindent
\textbf{Forward Pass.} In the forward pass, the problem is reduced to solving an ordinary differential equation (ODE) initial value problem. This is addressed by integrating the ODE from the initial time $t_0$ to the final time $t_u$, using an ODE solver as governed by the following equation: $\mathbf{Z}(t_u) = \mathbf{Z}(t_0) + \int_{t_0}^{t_u} S_\theta^{(a)}(\mathbf{z}(t), t)\, dt$. This expression implies that: $\mathbf{Z}(t_u) = \text{ODESolve}(\mathbf{z}(t_0), S_\theta^{(a)}, t_0, t_u)$, where the function \(\text{ODESolve}(\cdot)\) denotes the ODE solver. 

\noindent\textbf{Backpropagation.} The loss function can also be formulated as an integration problem, which can subsequently be solved using another ODE solver. This process involves reversing the time steps, starting from \( t_u \) and proceeding to \( t_0 \). The loss value can be determined using the following expression: $L(\textbf{z}(t_u)) = L\left( \textbf{z}(t_0) + \int_{t_0}^{t_u} S_\theta^{(a)} (\textbf{z}(t), t) \, dt \right)$, which can be equivalently represented as: $L(\textbf{z}(t_u)) = L\left(\text{ODESolve}(\textbf{z}(t_0), S_\theta^{(a)}, t_0, t_u )\right)$. The gradient is computed using the adjoint sensitivity method, which offers benefits such as constant memory cost and reduced numerical error. In particular, we employ the first-order adjoint method, as it has been shown to be more efficient, requiring fewer matrix computations compared to the Second-Order adjoint method, as demonstrated by \cite{norcliffe2020second}. We have previously transformed~\eqref{eq:important} into a system of first-order ordinary differential equations (ODEs), thereby enabling the direct application of the first-order adjoint method. 
We can compute the gradient by  $\frac{\partial L}{\partial \theta} = - \int_{t_u}^{t_0} \mathbf{a}(t)^T  \frac{\partial S_\theta^{(a)}(\mathbf{z}(t), t)}{\partial \theta} \, dt$, where $\mathbf{a}(t) = \frac{\partial L}{\partial z(t)}$ and $\frac{d\mathbf{a}(t)}{dt} = -\mathbf{a}(t) ^T \, \frac{\partial S_\theta^{(a)}(\mathbf{z}(t), t)}{\partial \mathbf{z}}$.
The integrals required for determining $\mathbf{z}$, $\mathbf{a}$ and $\frac{\partial L}{\partial \theta}$ can be efficiently computed in a single call of an ODE solver. This approach involves concatenating the original state, the adjoint state, and the additional partial derivatives into a unified vector, enabling simultaneous calculation.

\paragraph{Text-as-Image Augmentation.} \label{sec:tia}
CLIP establishes a robust shared image-text feature space, trained on large-scale image-text pairs, giving it powerful image-text correlation capabilities. Motivated by this unique characteristic, we propose using text as a form of data augmentation in this paper. Since text is easier and more efficient to obtain compared to images, this approach offers a practical and effective solution for augmenting data.

Specifically, we first build a prompt codebook, denoted as $\Psi \triangleq \{\psi_k\}_{k=1}^K$, where $\psi_k$ represents the prompt collection for class $k$. For each class, we select $M$ prompts from CLIP~\cite{radford2021learning} and CuPL~\cite{pratt2023does}. For example, for the class tiger shark,  a prompt from CLIP might be ``a photo of a {tiger shark}", while a prompt form CuPL could be ``A tiger shark typically has a dark blue or dark green upper body, with a light-colored underbelly". Given $N$ shots input images belonging to class $k$, we first exploit the image encoder $E_v$ to extract their image features $f_v$, and then calculate the average of these image features. We regard the average feature as the class prototype $f_v^k$. Then, we use the text encoder $E_t$ to generate the textual features $\{f_t^m\}_{m=1}^M$ from the prompts for class $k$ in $\Psi$. We calculate the cosine similarity between the image feature $f_v^k$ and the textual features by $\mathrm{sim}\left( f_v^k,f_t \right)=\frac{f_t\cdot f_v^k}{\Vert f_t \Vert  \Vert f_v^k \Vert}$. Next, we select the textual features which have the Top-$L$ similarity with the feature of the input image, using them as the augmentation features for class $k$. For example, for the $N$-shot setting, now we have $N+L$ features for training.

\subsection{Cross-Modal Few-Shot Learning}\label{sec:cross-modal}
The previous section introduced the Second-Order Neural ODE model, $\mathcal{S}^{(a)}_{\theta}(x)$. After the Text-as-Image Augmentation process, we obtain $N+L$ training features, denoted as $F_a \triangleq \{f_a^i\}_{i=1}^{N+L}$. It is important to note that these features are L2 normalized.
We next outline the training process for cross-modal few-shot learning. 
We begin by creating text samples by attaching the class label $y_k$ to a hand-crafted prompt such as $\pi =$ ``a photo of a". This process produces the text descriptions $t_k = \{\pi; y_k\}$ for each class $y_k$ in all $K$ classes.

During training, based on the features $F_a$, we learn a Second-Order Neural ODE model $\mathcal{S}^{(a)}_{\theta}(x)$ for optimizing the training features, and a cross-modal classifier $\phi$ for image classification. The classifier can be denoted as:
\begin{equation}
\label{eq:m_feature}
    \phi(x) = W^{\top}x,
\end{equation}
where $W$ represents the parameters of the cross-modal classifier $\phi$. These parameters are initialized using text features, with $w_{y_k}=E_t(t_k), \forall k \in [1, K]$. Here $w_{y_k}$ denotes the classification weight for class \(y_k\) in the parameter matrix \(W\).

Interestingly, this initialization strategy can be seen as a counterpart to prompt learning in CoOp~\cite{zhou2022learning}. Since these parameters are updated during training, they function similarly to prompt learning in CoOp. By initializing with text embeddings of prompts containing class labels, this approach eliminates the need to process data through the text encoder in each training iteration, making our method more efficient and effective.

The weights in $\phi$ and $\mathcal{S}^{(a)}_\theta$ can be updated by gradient descent with the following cross-entropy loss during training:

\begin{equation}
\begin{split}
\label{eq:celoss}
    \mathcal{L} _{CE} &= \sum_{i=1}^n{H\left( y_k,\phi \left( \mathcal{S}^{(a)}_{\theta} \left(f_a^i \right) \right) \right)} \\
    &= -\sum_{i=1}^n{\log \left( \frac{e^{w_{y_k}\cdot \mathcal{S}^{(a)}_{\theta} \left(f_a^i \right)}}{\sum\nolimits_{y'}{e^{w_{y'}\cdot \mathcal{S}^{(a)}_{\theta} \left(f_a^i \right)}}} \right)}.
\end{split}
\end{equation}

\paragraph{SONO Inference.} Given a test image $x_{test}$, we first utilize the image encoder $E_v$ to extract its image feature $f_v^{test}$, which is then input into the Second-Order Neural ODE Model $\mathcal{S}^{(a)}_{\theta}(x)$ to obtain the refined feature $\hat{f}_v^{test}$. The cross-modal classifier's prediction can be denoted as 
\begin{equation}
\label{eq-P_m}
    \mathcal{P}(y=y_k|x_{test})= w_{y_k} \cdot \hat{f}_v^{test},
\end{equation}
where $1\leqslant k\leqslant K$ is the class index. The final predicted label for the test image $x_{test}$ is then given by 
$\hat{y} = \underset{y'}{\mathrm{arg}\max}\,\mathcal{P}(y'|x_{test})$

\section{Experimental Results}

\subsection{Experimental Setting}
Following established protocols~\cite{zhou2022learning,zhang2022tip}, we conducted experiments on 11 benchmark \textbf{few-shot recognition} datasets. These datasets cover a wide range of tasks, including generic object classification, fine-grained object classification, remote sensing recognition, texture classification, scene recognition, and action recognition: ImageNet~\cite{recht2019imagenet}, Caltech101~\cite{fei2004learning}, OxfordPets~\cite{parkhi2012cats}, StanfordCars~\cite{krause20133d}, Flowers102~\cite{nilsback2008automated}, Food-101~\cite{bossard2014food}, FGVC Aircraft~\cite{maji2013fine}, DTD~\cite{cimpoi2014describing}, SUN397~\cite{xiao2010sun}, EuroSAT~\cite{helber2019eurosat}, and UCF101~\cite{soomro2012ucf101}. These datasets collectively serve as a robust benchmark for evaluating the few-shot learning capabilities of our model. Regarding \textbf{domain generalization}, we tested the model’s robustness to natural distribution shifts by training on a 16-shot ImageNet~\cite{deng2009imagenet} and evaluating it on four out-of-distribution variants: ImageNet-V2~\cite{recht2019imagenet}, ImageNet-Sketch~\cite{wang2019learning}, ImageNet-A~\cite{hendrycks2021natural}, and ImageNet-R~\cite{hendrycks2021many}.

\paragraph{Baselines.} We comprehensively compare our proposed SONO with state-of-the-art methods for vision-language models, including CLIP~\cite{radford2021learning}, CoOp~\cite{zhou2022learning}, PLOT~\cite{chen2023plot}, CLIP-Adapter~\cite{gao2021clip}, Tip-Adapter-F~\cite{zhang2022tip}, TaskRes~\cite{yu2023task}, and GraphAdapter~\cite{li2024graphadapter}. CoOp and PLOT are categorized as prompt learning methods, while TaskRes, CLIP-Adapter, Tip-Adapter-F, and GraphAdapter fall under adapter-style methods.

\paragraph{Implementation Details.}
We utilize CLIP~\cite{radford2021learning} as the base model, which consists of a ResNet-50 or ViT-B/16 image encoder and transformer text encoder.
During training, we freeze the weights of CLIP to leverage its pre-trained knowledge. Following CoOp~\cite{zhou2022learning}, we adopt the data preprocessing protocol from CLIP~\cite{radford2021learning}. we empirically set $M=50$ and $h = 10$. Our experiments use standard few-shot protocols with random selections of 1, 2, 4, 8, and 16 examples per class for training, followed by evaluation on the full test set. For domain generalization, we use the model trained on 16-shot ImageNet to evaluate its performance on four variants. We train our model for 20 epochs on ImageNet and 15 epochs on the other 10 datasets, using an initial learning rate of $1 \times 10^{-3}$. We optimize the model with the AdamW~\cite{kingma2015adam} optimizer and a cosine annealing scheduler. Our approach is parameter-efficient and lightweight, requiring only a single NVIDIA RTX 3090 GPU for training. Additional implementation details are provided in the Supplemental Materials.

\subsection{Performance Comparison}
\label{subsec:percom}

\begin{figure*}[!t]
\centering
\includegraphics[width=\textwidth]{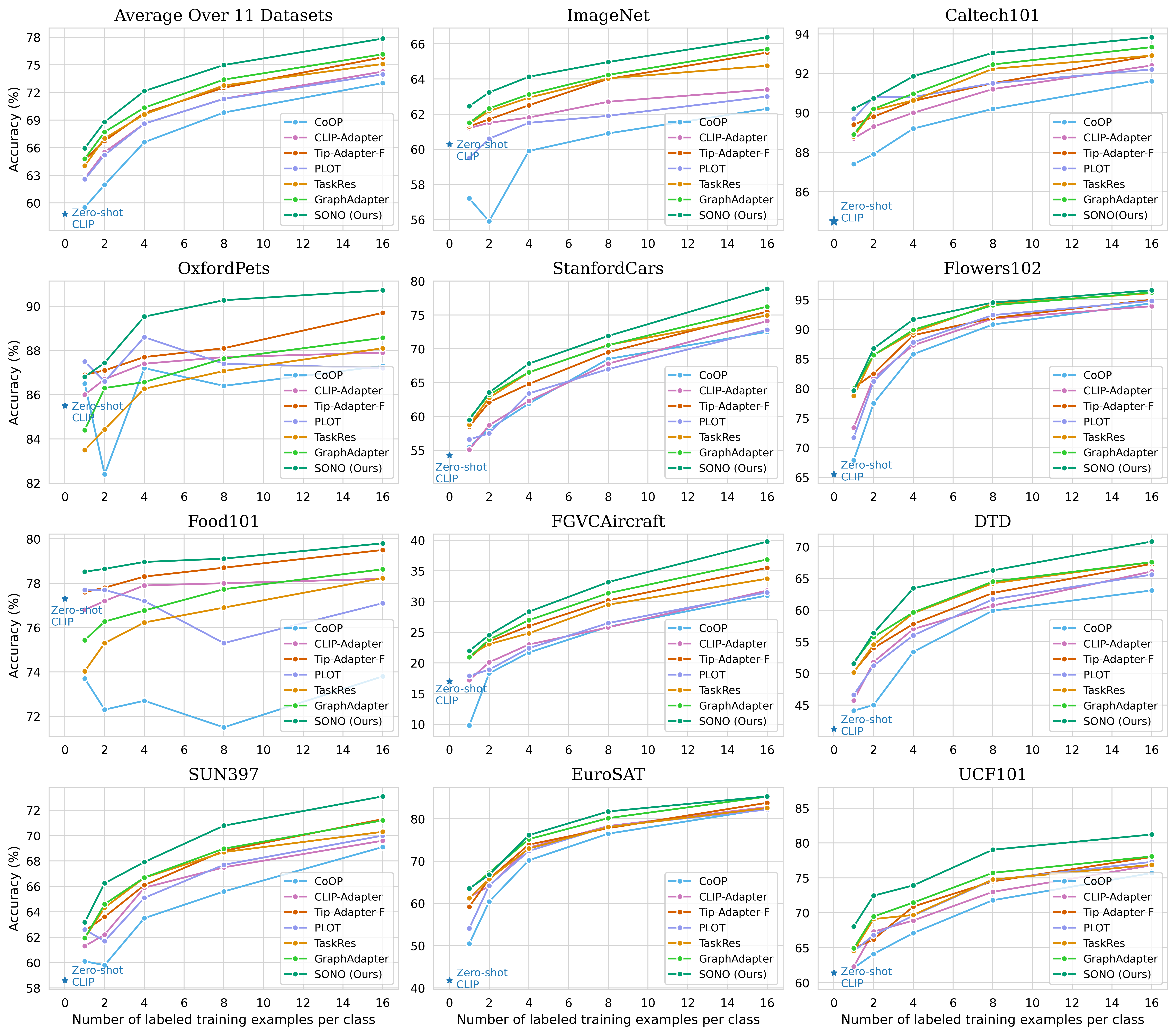}

\caption{\textbf{Classification Performance Comparison on Few-shot Learning}, {\em i.e.}, 1-/2-/4-/8-/16-shot, on 11 benchmark datasets. The top-left is the averaged accuracy over the 11 datasets.}
\label{fig:fewshot_results}

\end{figure*}

\paragraph{Few shot Recognition.}  
As shown in Figure~\ref{fig:fewshot_results}, we conduct a comprehensive evaluation of our proposed SONO across 11 datasets, each representing different tasks. We compare SONO with seven state-of-the-art methods, including both prompt learning and adapter-style approaches. The top-left sub-figure of Figure~\ref{fig:fewshot_results} illustrates the average accuracy results. Obviously, our proposed SONO method yields superior performance, consistently and substantially outperforming other methods from 1 to 16 shots, highlighting SONO's strong few-shot adaptation capability. In the 16-shot setting, SONO achieves an average performance of 77.86\%. This outperforms GraphAdapter~\cite{li2024graphadapter} by 1.63\% and Tip-Adapter~\cite{zhang2022tip} by 2.75\%. Moreover, the performance gain reaches 4.7\% compared to CoOp~\cite{zhou2022learning}. For the largest dataset, ImageNet, SONO outperforms the second-best method, GraphAdapter, by 0.67\%. On the most challenging datasets—FGVC Aircraft, DTD, and UCF101—our method achieves performance gains of up to 2.91\%, 3.26\%, and 3.14\%, respectively. The comprehensive results demonstrate the effectiveness and robust performance of SONO.

\begin{table}[]
\centering
\caption{Performance comparisons of domain generalization on two CLIP visual backbones. All models are trained on 16-shot ImageNet and tested on cross-domain datasets, including ImageNet-V2, -Sketch, -A, and -R.}
\resizebox{\linewidth}{!}{
\begin{tabular}{lccccccccc}
\toprule
\multirow{2}{*}{Method}            &  \multirow{2}{*}{Backbone} & Source &  & \multicolumn{5}{c}{Target}          \\  
\cmidrule(lr){3-3} \cmidrule(lr){5-9}
 &  & ImageNet & & -V2 & -Sketch & -A & -R & Average 
\\ \midrule
Zero-shot CLIP   & \multirow{6}{*}{\rotatebox{90}{ResNet-50}}  &  60.33  &      &   51.34          &   33.32              &   21.65         &   56.00         &  40.58       \\
Linear Probe CLIP &                             &   55.87 &     &      45.97       &  19.07                &   12.74        &  28.16          &    28.16     \\
CoOp            &                           &      62.95  &    &    55.11        & 32.74                 &     22.12      &   54.96         & 41.23           \\
TaskRes        &                             &     64.75 &   &    56.47           & 35.83                 &     22.80         &      60.70      &    43.95      \\
GraphAdapter  &                          &   \underline{65.70}    &     &   \underline{56.58}        &   \underline{35.89}              &        \underline{23.07} 
&    \textbf{60.86}       &  \underline{44.10}\\ 
Ours&                     &   \textbf{66.37} &       &   \textbf{57.83}  &\textbf{37.07}      &  \textbf{27.02}    
&    \underline{60.75}       &  \textbf{45.68}
\\ \midrule
Zero-shot CLIP    & \multirow{6}{*}{\rotatebox{90}{ViT-B/16}}   & 67.83  &       &  60.83         &   46.15                  & 47.77         &     73.96            & 57.18         \\
Linear Probe CLIP &                             &  65.85   &      &    56.26       &   34.77                  &    35.68        &  58.43        &    46.29        \\
CoOp            &                             &  71.92 &  &   64.18           & 46.71                      &     48.41       &    74.32           &  58.41       \\
TaskRes          &                             &      73.07  & &  65.30      &  49.13                        &   50.37    &            77.70 & 60.63           \\
GraphAdapter              &                             &   \underline{73.68}  &    &   \underline{66.60}        &             \underline{49.23}    &   \underline{50.75}       & \underline{77.73}          &  \underline{60.78}\\
Ours   &                             &   \textbf{74.92}  &    &   \textbf{67.33}        &             \textbf{51.15}    &   \textbf{52.96}       & \textbf{78.88}          & \textbf{62.58} \\
\bottomrule
\end{tabular}}
\label{tab:generalization}
\end{table}

\paragraph{Domain Generalization.} Table~\ref{tab:generalization} presents a comparison of the performance of our method against various baseline models. We provide classification results for the source domain (ImageNet~\cite{deng2009imagenet}) and several target domains: ImageNet-V2~\cite{recht2019imagenet}, ImageNet-Sketch~\cite{wang2019learning}, ImageNet-A~\cite{hendrycks2021natural}, and ImageNet-R~\cite{hendrycks2021many}. Additionally, we report the average accuracy across the out-of-distribution (OOD) datasets. Compared to the second-best method, GraphAdapter~\cite{li2024graphadapter}, our method improves the OOD average accuracy by 1.58\% on ResNet-50 and 1.80\% on ViT-B/16. Notably, we achieve a performance gain of up to 3.95\% on ImageNet-A. These results demonstrate that our SONO method exhibits exceptional robustness to distribution shifts.

\subsection{Ablation Studies}
We present an empirical analysis here.
Unless  specified, our experiments are conducted on the 16-shot ImageNet.

\paragraph{Contributions of Major Algorithm Components.}  In Table \ref{tab:ablation}, TAI stands for Text-Image Augmentation, SNM refers to the Second-Order NODEs, and the last row indicates a setup where only the classifier is retained. We conduct an ablation study by removing different components from our method to assess their impact. The first row shows the final performance of SONO during inference on the 16-shot dataset.
In the first ablation, removing TAI from SONO causes a performance drop of 0.71\%, emphasizing the importance of text-image augmentation. Next, we retain TAI but remove SNM, which optimizes input features for classification. This results in a 3.11\% decrease in performance, highlighting the significance of the Second-Order Neural ODE model. In the last row, with both TAI and SNM removed, accuracy drops to 62.35\%.
Overall, these results demonstrate that each component significantly enhances performance.

\begin{table}[t]
    \centering
    \caption{\textbf{Effectiveness of different algorithm components in SONO.}  TAI stands for Text-Image Augmentation, and SNM refers to the Second-Order Neural ODE model. The last row represents the setup where only the classifier is retained. We conducted experiments on ImageNet and reported the accuracy.}
    \label{tab:ablation}
    \setlength{\tabcolsep}{1mm}
    \scalebox{0.8}{
        \begin{tabular}{cc|ccccc} 
            \toprule
            TIA & SNM   & 1-shot & 2-shot & 4-shot & 8-shot &16-shot \\ 
            \midrule
            \cmark & \cmark &62.45 &63.23 &64.12 &64.96  & 66.37 \\
            \xmark & \cmark &60.81 &61.72 &62.85 &63.77 & 65.66 \\
            \cmark & \xmark &58.82 &59.19 &60.38 &61.57  & 63.26 \\
            \xmark & \xmark &57.65 &58.38 &59.66 &60.76 & 62.35 \\
            \bottomrule
        \end{tabular}
    }
\end{table}

\paragraph{Evaluation on Various Visual Backbones.}
Table \ref{table:backbones} summarizes the results on the 16-shot ImageNet \cite{deng2009imagenet} using various visual backbones, including ResNets~\cite{he2016deep} and ViTs~\cite{dosovitskiy2020image}. Our approach significantly improves performance, particularly when compared to zero-shot CLIP on the latest visual backbones. Additionally, our method consistently outperforms Tip-Adapter-F across all tested visual backbones.

\begin{table}[t!]
\caption{\textbf{Evaluation of various visual backbones}}
\label{table:backbones}
\centering
\resizebox{\linewidth}{!}{
\begin{tabular}{lcccc}
\toprule
Backbone& ResNet-50 & ResNet-101 & ViT-B/32 & ViT-B/16 \\
\midrule
Zero-shot CLIP         & 60.33     & 62.53      & 63.80    & 67.83    \\
CLIP-Adapter          & 63.59     & 65.39      & 66.19    & 71.13    \\
Tip-Adapter-F           & 65.51     & 68.56      & 68.65    & 73.69    \\

\textbf{SONO (Ours)}         & \textbf{66.37}     & \textbf{69.15}      & \textbf{69.71}    & \textbf{74.86}   \\
\bottomrule
\end{tabular}
}
\end{table}

\paragraph{Residual Ratio $\eta$.}
The hyperparameter $\eta$ controls the balance between the transformed features and the original features from the visual encoder when forming the final visual features. A larger $\eta$ indicates greater reliance on the transformed features. As shown in Table \ref{tb:hyper}, classification accuracy improves as $\eta$ increases from 0.0, peaking at 66.37\% when $\eta = 0.6$. This suggests that the transformed features generated by the Second-Order Neural ODE significantly enhance the final prediction.

\begin{figure}[t!]
\setlength{\abovecaptionskip}{0.2cm}
    \centering
    \includegraphics[width = 0.9\linewidth]{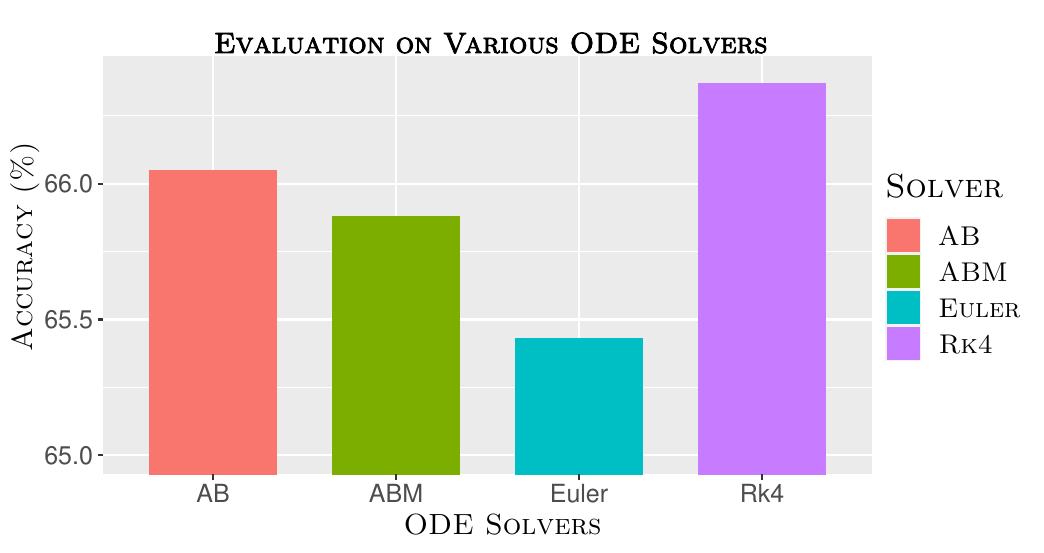}
    \caption{Ablation results on various ODE solvers: Fourth-Order Runge-Kutta (RK4), Euler, Explicit Adams-Bashforth (AB), and Implicit Adams-Bashforth-Moulton (ABM) methods.}
    \label{fig:solver}
\end{figure}

\begin{table}[t!]
\caption{\textbf{Sensitivity of hyperparameters}. All the results are reported on a $16$-shot setting on ImageNet.}
\centering
\begin{adjustbox}{width=0.8\linewidth}
	\begin{tabular}{c|cccccc}
	\toprule
		\multicolumn{7}{c}{Sensitivity of Hyperparameters} \\ 
		\midrule
		$\eta$  &0.0 &0.2 &0.4&\textbf{0.6} &0.8 &1.0 \\  
	    \cmidrule(lr){1-1}\cmidrule(lr){2-7}
        \cmidrule(lr){1-7}
		 Acc. &64.06  & 65.51  &66.20 &\textbf{66.37} &66.12 &65.93\\ 
	\bottomrule
	\end{tabular}
\end{adjustbox}
\label{tb:hyper}

\end{table}

\begin{table}[t!]
\centering
\caption{\textbf{Efficiency comparisons on 16-shot ImageNet}. We report the results using only a  NVIDIA RTX 3090 GPU.}
\label{table:efficiency}
\resizebox{\linewidth}{!}{
\begin{tabular}{l|cccccc}
\toprule
Method   & Epochs    & Training & GFLOPs  & Param. & Acc. & Gain \\
\midrule
CoOp  & 200   & 15 h & $>$10 & \textbf{0.01M}   & 62.95 & -\\
CLIP-Adapter  & 200   & 50 min & 0.004 & 0.52M   & 63.59 & +0.64\\
Tip-Adapter-F  & \textbf{20}   & 5 min & 0.030 & 16.3M   & 65.51 & +2.56\\

\textbf{SONO(Ours)} & \textbf{20}   & \textbf{3.5 min} & \textbf{0.010} & 1.54M   & \textbf{66.37} & \textbf{+3.42} \\
\bottomrule
\end{tabular}
}
\end{table}

\paragraph{Evaluation on Various ODE Solvers.}
Figure \ref{fig:solver} presents the results of different ODE solvers on the 16-shot ImageNet dataset. Our qualitative analysis aligns with the empirical findings, indicating that the RK4 outperforms the Euler, AB, and ABM methods.

\paragraph{Efficiency Comparison.} 
To demonstrate the exceptional fine-tuning efficiency of our method, we compare it against other state-of-the-art approaches, considering training epochs, training time, computational cost, and the number of parameters. The detailed results are presented in Table~\ref{table:efficiency}. Our SONO method reaches an accuracy of 66.37\% on 16-shot ImageNet in just 3.5 minutes, using 1.54 million parameters. In comparison, CoOp needs about 15 hours of training to achieve 62.95\% accuracy, while Tip-Adapter-F takes 5 minutes and requires 16.3 million parameters to achieve 65.51\% accuracy. Additional details 
are provided in the Supplemental Materials.

\label{sec:experiments}

\section{Conclusion}\label{sec:conclusion}
We introduce SONO, a method for cross-modal few-shot learning, addressing the critical challenge of overfitting with limited data. By leveraging Second-Order NODEs with an efficient cross-modal classifier, SONO significantly enhances feature optimization and generalization capabilities. Utilizing text-based image augmentation further strengthens the training process, optimizing the model's performance across diverse datasets. Our experiments demonstrate that SONO not only surpasses existing few-shot learning models but also shows proficiency in domain generalization.

\section{Acknowledgments}
CWC is supported by the  Swiss National Science Foundation (SNSF)  under grant number 20HW-1\_220785. YZ, ZH were supported by the National Natural Science Foundation of China (No. 62331014) and 2021JC02X103. CBS acknowledges support from the Philip Leverhulme Prize, the Royal Society Wolfson Fellowship, the EPSRC advanced career fellowship EP/V029428/1, EPSRC grants EP/S026045/1 and EP/T003553/1, EP/N014588/1, EP/T017961/1, the Wellcome Innovator Awards 215733/Z/19/Z and 221633/Z/20/Z, CCMI and the Alan Turing Institute. AIAR gratefully acknowledges support from the Yau Mathematical Sciences Center, Tsinghua University.

\bigskip

\bibliography{aaai25}

\end{document}